\documentclass[sigconf]{acmart}

\usepackage{multirow}  
\usepackage[ruled,vlined]{algorithm2e}
\usepackage{algorithmic}
\usepackage{color}

\newcommand{\mname}{MMF-AC}

\AtBeginDocument{%
  \providecommand\BibTeX{{%
    \normalfont B\kern-0.5em{\scshape i\kern-0.25em b}\kern-0.8em\TeX}}}

\copyrightyear{2022}
\acmYear{2022}
\setcopyright{rightsretained}
\acmConference[KDD '22]{Proceedings of the 28th ACM SIGKDD Conference on Knowledge Discovery and Data Mining}{August 14--18, 2022}{Washington, DC, USA.}
\acmBooktitle{Proceedings of the 28th ACM SIGKDD Conference on Knowledge Discovery and Data Mining (KDD '22), August 14--18, 2022, Washington, DC, USA}
\acmDOI{10.1145/3534678.3542672}
\acmISBN{978-1-4503-9385-0/22/08}
\begin{document}

\title{Scalable Online Disease Diagnosis via Multi-Model-Fused Actor-Critic Reinforcement Learning}

\author{Weijie He}
\affiliation{%
  \institution{Department of Computer Science and Technology \&
  Institute of Artificial Intelligence \& BNRist, \\ Tsinghua University}
  \city{Beijing}
  \country{China}
}
\email{hwj19@mails.tsinghua.edu.cn}

\author{Ting Chen}
\authornote{Corresponding author.}
\affiliation{%
  \institution{Department of Computer Science and Technology \&
  Institute of Artificial Intelligence \& BNRist, \\ Tsinghua University}
  \city{Beijing}
  \country{China}
}
\email{tingchen@tsinghua.edu.cn}

\begin{abstract}
For those seeking healthcare advice online, AI based dialogue agents capable of interacting with patients to perform automatic disease diagnosis are a viable option. This application necessitates efficient inquiry of relevant disease symptoms in order to make accurate diagnosis recommendations. This can be formulated as a problem of sequential feature (symptom) selection and classification for which reinforcement learning (RL) approaches have been proposed as a natural solution. They perform well when the feature space is small, that is, the number of symptoms and diagnosable disease categories is limited, but they frequently fail in assignments with a large number of features. To address this challenge, we propose a Multi-Model-Fused Actor-Critic (\mname) RL framework that consists of a generative actor network and a diagnostic critic network. The actor incorporates a Variational AutoEncoder (VAE) to model the uncertainty induced by partial observations of features, thereby facilitating in making appropriate inquiries. In the critic network, a supervised diagnosis model for disease predictions is involved to precisely estimate the state-value function. Furthermore, inspired by the medical concept of differential diagnosis, we combine the generative and diagnosis models to create a novel reward shaping mechanism to address the sparse reward problem in large search spaces. We conduct extensive experiments on both synthetic and real-world datasets for empirical evaluations. The results demonstrate that our approach outperforms state-of-the-art methods in terms of diagnostic accuracy and interaction efficiency while also being more effectively scalable to large search spaces. Besides, our method is adaptable to both categorical and continuous features, making it ideal for online applications.
\end{abstract}




\maketitle

\begin{figure}
  \centering
  \includegraphics[width=0.9\linewidth]{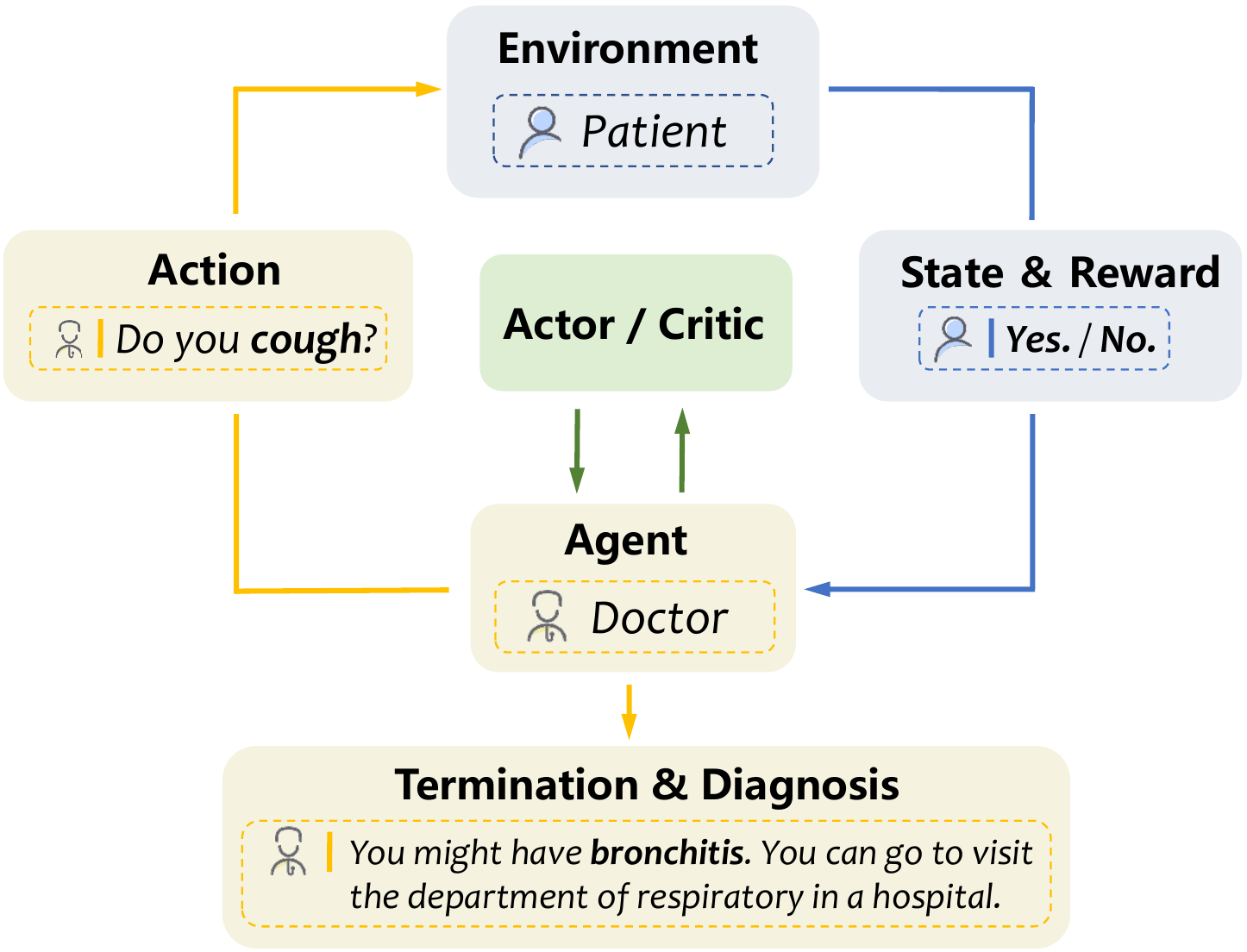}
  \caption{Logic of reinforcement learning in an automatic disease diagnosis process based on dialogue.}
  \Description{The user goes for an online disease diagnosis, representing the environment. The agent inquires whether the user has a cough as an action. The user would answer yes or no. Finally, the agent terminates the interaction and diagnoses that the user might have bronchitis and suggests that he should visit the department of respiratory.}
  \label{fig:example}
\end{figure}

\section{Introduction}

Recent advances in machine learning have facilitated the development of an alternative "AI doctor", that is, the ability to develop a disease diagnosis model empirically from growing collections of clinical observations. Indeed, automatic disease diagnosis has the potential to benefit both doctors and patients in today's medical system. Primary care physicians, for example, who are responsible for coordinating patient care among specialists and other levels of care, regularly encounter a variety of illnesses that should be assigned to different departments. In this situation, they may use the automatic disease diagnosis tool to guide what healthcare treatments patients require. Furthermore, the rapid growth of the mobile web has increased customer access to online self-diagnosis. Numerous self-diagnosis tools, such as the well-known symptom checkers from Mayo Clinic \cite{mayo} and WebMD \cite{webmd}, have been developed to address these needs. Adults are increasingly turning to these online services for advice on their primary concerns prior to deciding whether to see the primary care physicians or which specialists to visit \cite{fox2013health}. It is important that online disease diagnosis technologies are not a substitute for medical investigations by a doctor. Physical examinations and diagnostic tests such as X-rays and laboratory tests, which are crucial in the diagnosis of many diseases, could not be done online. However, there is indeed a genuine need for an efficient and accurate online "AI doctor" in healthcare \cite{cross2021search}. 

An "AI doctor" would imitate the diagnosis procedure of human doctors by gathering patients' symptoms through dialogues until it could confidently make a diagnosis. 
Naturally, more information regarding possible symptoms could lead to a more accurate diagnosis; for example, most existing questionnaire systems inquire about a considerable number of symptoms from patients \cite{shim2018joint,lewenberg2017knowing}. Unfortunately, patients may lose patience during the inquiry process due to the inefficiency of such systems. To overcome this problem, dialogue-based diagnosis agents have been proposed to reduce the number of inquiries while achieving high diagnostic accuracy \cite{tang2016inquire}. 

Dialogue-based disease diagnosis can be formulated as a cost-sensitive sequential feature selection and classification task in machine learning. Inquiring a symptom to obtain its value is comparable to selecting a feature that comes with a cost. Reinforcement learning (RL) approaches have demonstrated good performance in sequential feature selection and classification tasks with a small number of features \cite{tang2016inquire, kachuee2018opportunistic, kao2018context, peng2018refuel, xia2020generative, liao2020task, lin2020towards, liu2021dialogue}. Figure~\ref{fig:example} illustrates how RL works in the process of automatic disease diagnosis. However, RL frequently struggles in dealing with large feature spaces. Without a well-defined exploration strategy, an agent may fail to obtain useful rewards for learning in a large action space where most actions are unrelated to the target.

To address the challenge, we propose a scalable Multi-Model-Fused Actor-Critic RL framework (\mname) that integrates a generative model and a supervised classification model into an actor-critic RL algorithm. The actor-critic algorithm consists of two networks, the actor and the critic, working together to solve the online disease diagnosis problem. At a high level, the actor decides which action should be taken next, and the critic informs the actor how good the action is and how it should adjust. Based on the observed symptoms, the actor of the RL agent must decide whether to continue inquiring (take action) or make a diagnosis (terminate), as well as which subsequent symptom to check if the decision to continue inquiring is made. If the actor decides to terminate the interaction, the supervised classification model is invoked to make a disease diagnosis. Naturally, an unobserved symptom with a high conditional probability given the observed symptoms should be inquired about next since the patient is likely to have it. Hence to speed up convergence during training, we develop a novel actor network by leveraging the Variational AutoEncoder (VAE) \cite{kingma2013auto} that models the conditional probability distributions between observed and unobserved symptoms. In our customized critic network, we make use of the intermediate diagnostic confidence from the supervised classification model to help criticize the actions made by the actor since it would reinforce the agent to inquire about the proper symptoms that contribute to the diagnosis. 

Inspired by the concept of differential diagnosis in healthcare, we create a novel differential reward shaping mechanism. Differential diagnosis is the process of distinguishing a particular disease from others that share similar symptoms \cite{LAMBA20211, sajda2006machine}.  To narrow down the list of possible diseases, physicians typically inquire about specific symptoms, obtain a medical history, conduct a physical examination, or order diagnostic laboratory or imaging tests. This diagnostic process can be considered as inquiring key features in order to confirm or rule out potential diseases. To implement this concept within the context of RL, the RL agent should hypothesize a list of possible diseases based on observed symptoms and then use the reward function to prioritize key symptoms for differential diagnosis. However, estimating the reward function can be very challenging when the number of symptoms is large. Many existing RL algorithms \cite{liao2020task, peng2018refuel, liu2021dialogue} take a greedy approach to constructing reward functions by considering only "positive symptoms", that is, those that the patient actually experiences, while ignoring all other “negative symptoms”, those that the patient does not have. Because positive symptoms account for a very small proportion of all possible symptoms, the greedy mechanism results in the sparse reward problem, which poses a scalability challenge for RL algorithms. Besides, this greedy mechanism would direct the agent to investigate symptoms that frequently co-occur while ignoring negative symptoms that are critical for differential diagnosis. In comparison, our differential reward shaping mechanism accounts for both the rate of co-occurrence of symptoms and the differential effect by using generative and diagnosis models.

Unlike the majority of previous RL baselines, which have been proven to be effective exclusively on either synthetic or real-world datasets, our approach is effective in both types of datasets. The results demonstrate that our method achieves the state-of-the-art performance on real-world datasets and is more effectively scalable to large search spaces on synthetic datasets, where most RL methods fail. Furthermore, we undertake an evaluation on a real-world diabetes dataset, showing that our approach, unlike conventional baselines, can accommodate continuous feature in addition to categorical symptoms. 

Our contributions are highlighted in the following:
\begin{itemize}
\item We propose a Multi-Model-Fused Actor-Critic RL framework (\mname) that achieves state-of-the-art performance across various settings. Our approach is suitable for online applications due to its scalability and generalizability;

\item Technically, we leverage the characteristics of the VAE and the supervised classification model to enhance the agent with a generative actor and a diagnostic critic network;

\item We present a novel differential reward shaping mechanism to address the sparse reward problem in large search spaces.
\end{itemize}

\section{Related Work}

\subsection{Dialogue-Based Disease Diagnosis}

Deep reinforcement learning methods have been extensively utilized to address the problem of automatic disease diagnosis based on dialogue. \citet{tang2016inquire} first formulated the inquiry and diagnosis process as a Markov decision process and then employed reinforcement learning (RL) in a simulated environment. \citet{kao2018context} and \citet{peng2018refuel} showed the feasibility of achieving competitive results in medium search spaces. \citet{kachuee2018opportunistic} suggested a deep Q-learning method with model uncertainty variations as the reward function. In the task of sequential feature acquisition with a specified cost function, \citet{janisch2019classification, janisch2020classification} found that RL methods outperformed tree-based methods. \citet{xia2020generative} developed an RL agent for inquiry and diagnosis tasks using the Generative Adversarial Network (GAN) and the policy gradient approach. The RL agent performed well on two public datasets with four and five diseases, respectively. However, in most real-world settings, such a tiny number of diseases is impractical. \citet{lin2020towards} proposed a RL model combining an inquiry branch implemented through a Q network with a diagnosis branch involving a generative sampler. It was tested only on the two small datasets described previously. \citet{liao2020task} developed a  hierarchical RL architecture with a master for worker appointment, several workers for symptom inquiry, and a distinct worker for disease screening. \citet{liu2021dialogue} improved its convergence employing a pretraining strategy, but they were limited to manage up to 90 different diseases.  

There are also a number of non-RL approaches to inquiry-based disease diagnosis. Entropy functions have been used to identify symptoms based on the principle of information gain in a diverse family of Bayesian inference and tree-based approaches \cite{ kononenko1993inductive,kohavi1996scaling,kononenko2001machine,xu2013cost,zakim2008underutilization,nan2016pruning}. 
\citet{wang2021online} developed a hierarchical lookup framework based on symptom embedding and disease-specific graph representation learning. Due to the intractable nature of maximization of global  information gain or global sensitivity, these methods frequently rely on greedy or approximation algorithms that result in low accuracy. Recently, EDDI \cite{ma2019eddi}, a framework for active feature acquisition, was proposed as a method for selecting the next feature by maximizing a defined information-theoretic reward over all features. It employs a Partial Variational AutoEncoder (VAE) to deal with partially observed data when inquiring. The computational cost of estimating the reward using Monte Carlo sampling, on the other hand, is too costly to handle a large number of symptoms. \citet{he2022bsoda} adopted EDDI into a framework for online disease diagnosis. They proposed a bipartite scalable framework and demonstrated its state-of-the-art performance on a variety of datasets, as well as its ability to deal with very large feature sets. Although several novel strategies have been designed to reduce the computational cost of the reward, the response time is still too slow to support online applications.

\subsection{Actor-Critic Reinforcement Learning}

Q learning \cite{watkins1992q} and Deep Q networks \cite{mnih2013playing} are examples of value-based reinforcement learning methods attempting to compute the optimal action-value function. While value-based methods are efficient and stable, policy-based methods, such as policy gradient techniques \cite{lillicrap2015continuous}, which conduct gradient ascent optimization on the performance objective to determine the best policy, are certain to converge and easy to adapt to various contexts \cite{nachum2017bridging}. Taking advantage of both types of methodologies, the actor-critic algorithm learns an actor network computing the probabilities of actions and a critic network computing the state-value function. In particular, the actor-critic approach optimizes an advantage function rather than the state-value function. Many variants of actor-critic have emerged recently. For instance, \citet{mnih2016asynchronous} developed an asynchronous variant of actor-critic, called asynchronous advantage actor-critic (A3C). \citet{schulman2015trust} proposed a practical variant, Trust Region Policy Optimization (TRPO), by making several approximations to the theoretically-justified scheme. Proximal Policy Optimization (PPO) \cite{schulman2017proximal} is an approach based on TRPO for learning a stochastic policy by minimizing a clipped surrogate goal. \citet{haarnoja2018soft} also suggested an off-policy maximum entropy actor-critic method, commonly known as Soft Actor-Critic (SAC), enabling sample-efficient learning while maintaining stability.

\section{Methodology}

In this section, we first present the detailed formulation of the problem and then introduce the structure of the generative actor and diagnostic critic, as shown in Figure~\ref{fig:ar}. Next, we elaborate on the differential reward shaping mechanism. Finally, we give an overview of the proposed Multi-Model-Fused Actor-Critic algorithm (\mname), as shown in Figure~\ref{fig:flowchart}.

\begin{table}
  \caption{Notations.}
  \label{tab:not}
  \begin{tabular}{cc}
    \toprule
    Notation & Description\\
    \midrule
    $x$ & Presence or absence of a symptom.\\
    $\mathbf{x}_O/\mathbf{x}_U$ & Status of observed/unobserved symptoms.\\
    $\mathbf{y}$ & Presences or absences of diseases.\\
    $N_S/N_D$ & Number of total symptoms/diseases in a dataset.\\
    $N_O$ & Number of observed symptoms.\\
    $N_T$ & Number of maximum inquiries.\\
    $a_t/\mathbf{s}_t$ & Action/State at timestep $t$.\\
    $\pi_\theta(a_t|\mathbf{s}_t)$ & Optimized policy with parameters $\theta$.\\
    $R(\mathbf{s}_t,\mathbf{s}_{t+1})$ & Reward of the transition from $\mathbf{s}_t$ to $\mathbf{s}_{t+1}$.\\
    $V_\pi(\mathbf{s}_t)$ & State-value function of state $\mathbf{s}_t$ given policy $\pi$.\\
    $\mathcal{D}(\cdot)$ & Supervised diagnosis model.\\
  \bottomrule
\end{tabular}
\end{table}

\subsection{Problem Formulation}
\label{sec:bg}

In this paper, we aim to learn a dialogue-based agent for automatic disease diagnosis. The input data consists of a set of explicit symptoms that the patient self-reports prior to the doctor's inquiry and a set of implicit symptoms that the patient experiences but are to be acquired through the dialogue. The output is a disease label diagnosed by the doctor. The objective is to efficiently  acquire the patient's symptom conditions in order to establish an accurate diagnosis. This can be described as a Markov Decision Process (MDP) \cite{bellman1957markovian}, which can be characterized by a tuple $\mathcal{M}(\mathcal{S},\mathcal{A},\mathcal{T},R,\gamma)$. $\mathcal{S}$ (state) is a set of vectors that represents the current state of the dialogue. $\mathcal{A}$ (action) specifies the inquiry through a collection of available actions, including inquiring about a specific symptom or terminating the dialogue in order to make a diagnosis in our setting. As a result, the objective is to determine the optimal conversational policy $\pi_\theta:\mathcal{S}\rightarrow\mathcal{A}$ that is composed of a series of actions. At each step $t$, the policy $\pi_\theta(a_t|\mathbf{s}_t)$ seek to determine the action $a_t\in\mathcal{A}$ based on the current state $\mathbf{s}_t\in\mathcal{S}$. The transition function $\mathcal{T}$ is the probability distribution over the next state $\mathbf{s}_{t+1}$ given current state $\mathbf{s}_t$ and action $a_t$, or $p(\mathbf{s}_{t+1}|\mathbf{s}_t,a_t)$. $R$ is a reward function that evaluates the gain of the transition $(\mathbf{s}_t,a_t,\mathbf{s}_{t+1})$, and $\gamma$ is the discount factor for accumulative rewards at each step. The state-value function that RL approaches strive to maximize is the expected sum of discounted rewards given the state $\mathbf{s}_t$ corresponding to policy $\pi_\theta$:
\begin{equation}
\label{eq:v}
    V_\pi(\mathbf{s}_t)=\mathbb{E}[R_t+\gamma R_{t+1}+\gamma^2 R_{t+2}+...|\mathbf{s}_t].
\end{equation}
In our setting, when the RL agent terminates the interaction, a supervised classification model is invoked to produce a disease diagnosis. Let $N_O$ denote the number of observed symptoms and $\mathbf{x}_O\in\{0,1\}^{N_O}$ denote the presence (1) or absence (0) of observed symptoms. Similarly, let $N_D$ denote the number of distinct diseases and $\mathbf{y}\in\{0,1\}^{N_D}$ denote a one-hot categorical vector with each dimension indicating whether a specific disease is present (1) or not (0). The diagnosis model is then defined as a mapping $\mathcal{D}$ between the observed symptoms $\mathbf{x}_O$ and the disease probability distribution $\hat{\mathbf{y}}\in(0,1)^{N_D}$. $\mathcal{D}$ can be any supervised classification model that outputs prediction probabilities. For example, HiePPO \cite{liu2021dialogue} applied LightGBM \cite{ke2017lightgbm} method, while BSODA \cite{he2022bsoda} developed a supervised self-attention model to embed relationships between all features, including diseases and symptoms, and perform diagnosis using the disease embeddings. As BSODA shows great performance on a variety of datasets, we adopt its diagnosis model as $\mathcal{D}$. Please refer details to BSODA \cite{he2022bsoda}. Table~\ref{tab:not} summarizes some of the notations.

\subsection{Generative Actor}
\label{sec:ga}
\begin{figure}
  \centering
  \includegraphics[width=\linewidth]{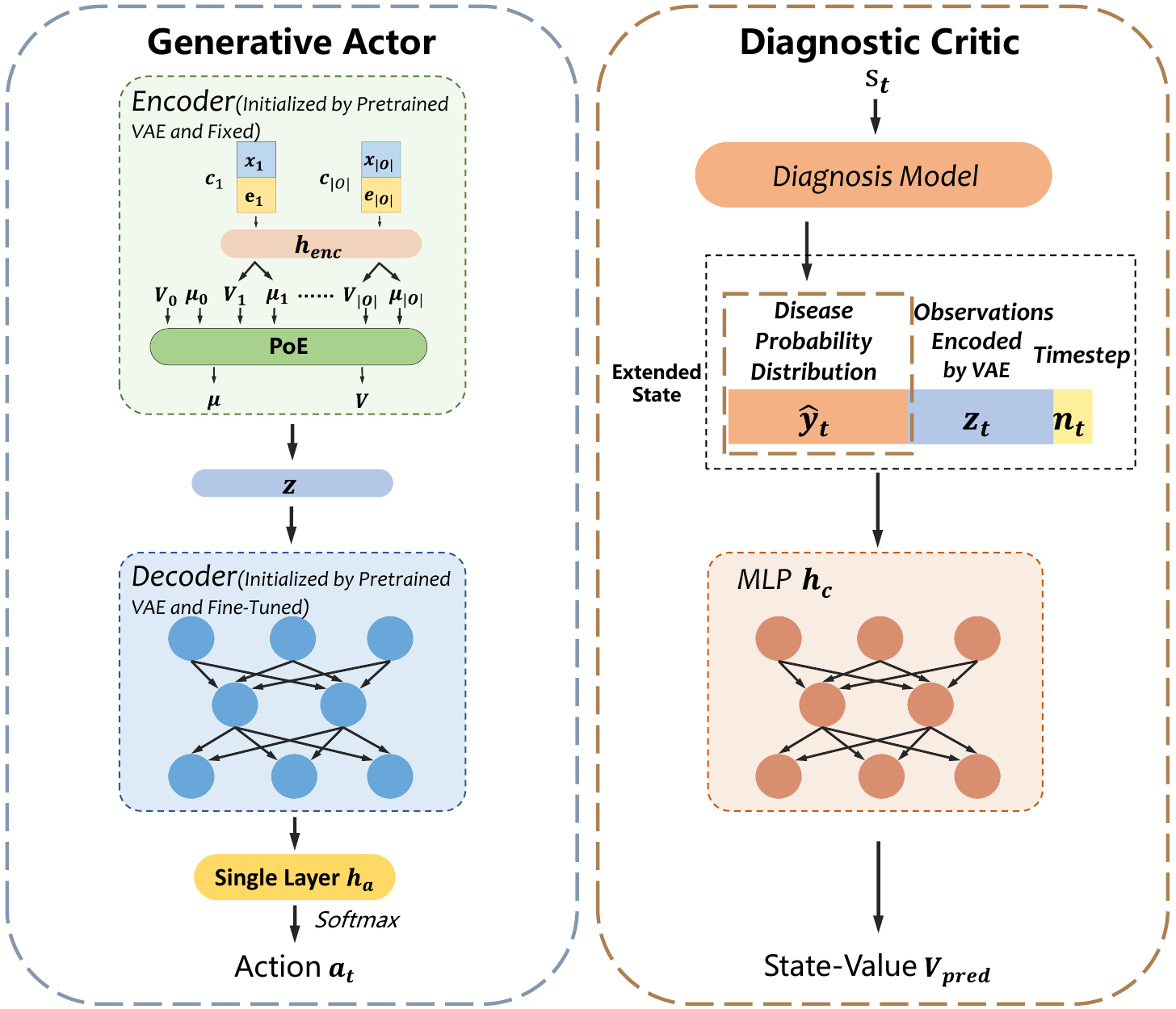}
  \caption{Detailed structures of the generative actor and diagnostic critic of \mname.}
  \Description{We imitate the VAE to build our generative actor network structure and then map the decoder output to the action space using a single layer network as $h_a(\cdot)$ with the softmax activation function. In our diagnostic critic, we extend the state by three parts: $[\hat{y}_t,\mathbf{z}_t,n_t]$. $\hat{y}_t=\mathcal{D}(\mathbf{s}_t)\in(0,1)^{N_D}$ is the output disease probability distributions of the supervised diagnosis model. $\mathbf{z}_t$ is the latent embedding of the VAE encoder receiving $\mathbf{x}_O$ as input. $n_t=t/N_T$ means the number of inquiries that have been made, which is divided by the pre-defined maximum number of inquiries $N_T$. As a result, we use an MLP as $h_c(\cdot)$ to map the informative state vector to predicted state-value $V_{pred}$.}
  \label{fig:ar}
\end{figure}

The left panel of Figure~\ref{fig:ar} displays the structure of the generative actor. Action $a_t$ in our framework is either inquiring about a specific symptom or terminating the dialogue, and it has up to $N_S+1$ possible values where $N_S$ is the total number of symptoms. Naturally, an unobserved symptom $x$ with a high conditional probability $p(x|\mathbf{x}_O)$ should be investigated next since the patient is likely to have this symptom. We model the conditional distributions between observed symptoms $\mathbf{x}_O$ and unobserved symptoms $\mathbf{x}_U$ in the following. A VAE defines a generative model of the form $p(\mathbf{x}, \mathbf{z})=\prod_i p_\Theta(x_i|\mathbf{z})p(\mathbf{z})$ where the data $\mathbf{x}$ is generated from latent variables $\mathbf{z}$. $p(\mathbf{z})$ is a prior, e.g., spherical Gaussian, and $p_\Theta(\mathbf{x}|\mathbf{z})$ is presented as a neural network decoder with parameters $\Theta$ specifying a simple likelihood, e.g., Bernoulli. A VAE produces a variational approximation of the posterior $q_\Phi(\mathbf{z}|\mathbf{x})$ by using another neural network with parameters $\Phi$ as an encoder. The model can be trained by optimizing a lower bound on the evidence (ELBO):
\begin{equation}
\label{eq:ELBO}
    \mathbb{E}_{q_\Phi(\mathbf{z}|\mathbf{x})}[\log p_\Theta(\mathbf{x}|\mathbf{z})]-\beta\cdot D_\textrm{KL}[q_\Phi(\mathbf{z}|\mathbf{x})\,\|\, p(\mathbf{z})],
\end{equation}
where $\beta$ is the weight to balance the two terms in the expression and $D_{\textrm{KL}}$ is the Kullback-Leibler (KL) divergence between two distributions. 
Then, $p(\mathbf{x}_U|\mathbf{x}_O)$ can be obtained by first sampling $\hat{\mathbf{z}}\sim q_\Phi(\mathbf{z}|\mathbf{x}_O)$ from the VAE encoder, followed by $p_\Theta(\mathbf{x}_U|\hat{\mathbf{z}})$ from the VAE decoder. The most challenging aspect is that, during the inquiry process, observed symptoms are usually incomplete, resulting in up to $O(2^{N_S})$ distinct $q(\mathbf{z}|\mathbf{x}_O)$. It is impossible to train an independent encoder network for every distribution. To efficiently handle the partial observations, we propose to use the Product-of-Experts (PoE) encoder \cite{wu2018multimodal}, which takes advantage of the nature of VAE's latent variables $\mathbf{z}$. Specifically, let $\mathbf{e}_i$ denote the embedding vector of the $i$-th feature $x_i$ and $\mathbf{c}_i=[x_i, \mathbf{e}_i]$ denote the concatenated input carrying information of the $i$-th observed feature. Assuming conditional independence among features, the approximated posterior, including a prior Gaussian expert $p(\mathbf{z})$ with mean vector $\mu_0$ and variance vector $V_0$, is given by
\begin{equation}
\label{eq:mvae}
    q_\Phi(\mathbf{z}|\mathbf{x}_O)\propto p(\mathbf{z})\prod q_\Phi(\mathbf{z}|x_i),
\end{equation}
where $q_\Phi(\mathbf{z}|x_i)$ is an inference network denoting the expert associated with the $i$-th observed feature $x_i$. Then we use a Multiple Layer Perceptron (MLP) as $h_{enc}(\cdot)$ to map the input $\mathbf{c}_i$ to a Gaussian distribution in latent space with mean vector $\mu_i$ and variance vector $V_i$. $h_{enc}(\cdot)$ shares parameters across all features. Because a product of Gaussian experts is itself Gaussian \cite{cao2014generalized}, we can quickly compute variance $V=(V_0^{-1}+\sum V_i^{-1})^{-1}$ and mean $\mu=(\mu_0 V_0^{-1}+\sum\mu_i V_i^{-1})V$ for any $q_\Phi(\mathbf{z}|\mathbf{x}_O)$ in Eqn.~\ref{eq:mvae}. The decoder, $p_\Theta(\mathbf{x}|\mathbf{z})$, is given by a product of Bernoulli distributions, the probabilities of which are specified by an MLP that accepts $\mathbf{z}$ as input. 

To develop our generative actor network, we imitate the VAE structure and then use a single layer network $h_a(\cdot)$ with a softmax activation function to map the decoder output to the action space. During training, we pretrain a standard VAE and then use its parameters to initialize the actor's encoder and decoder. Only the actor's decoder will be fine-tuned. In the early stages of training, the actor can use the pretrained parameters to investigate significant symptoms and improve the learning efficiency. The pretrained VAE will also be used in the next sections.

\subsection{Diagnostic Critic}
\label{sec:diag}

The right panel of Figure~\ref{fig:ar} displays the structure of the diagnostic critic. The critic's objective is to precisely compute the state-value function $V_\pi(\mathbf{s}_t)$ in Eqn.~\ref{eq:v} in order to optimize the policy $\pi_\theta$. At timestep $t$, the state $\mathbf{s}_t\in \{1,0,-1\}^{N_S}$ represents the current symptom information, with 0 indicating a symptom that has not been investigated, 1 indicating a \textbf{positive symptom} or that the patient has experienced the symptom, and -1 indicating a \textbf{negative symptom} or that the patient did not experience the symptom. However, the diagnostic confidence from the diagnosis model is also significant to evaluate the current state, which would reinforce the agent to inquire about the proper symptoms that contribute to the diagnosis. The intermediate diagnostic result and the current timestep $t$ can also help determine when to terminate the interaction and produce a confident final diagnosis. Furthermore, to handle arbitrary partially observations in large feature spaces, a powerful encoder is required. Hence in our diagnostic critic, we extend the state by three parts: $[\hat{y}_t,\mathbf{z}_t,n_t]$ where $\hat{y}_t=\mathcal{D}(\mathbf{s}_t)\in(0,1)^{N_D}$ is the output disease probability distribution from the supervised diagnosis model, $\mathbf{z}_t$ is the latent embedding of the aforementioned pretrained VAE encoder receiving $\mathbf{x}_O$ as input, and $n_t=t/N_T$ is the ratio of the number of inquiries having been made over $N_T$, a pre-defined maximum number of inquiries. As a result, we use an MLP, $h_c(\cdot)$, to map the informative state vector to predicted state-value $V_{pred}$.

\subsection{Differential Reward Shaping}
\label{sec:r}
Differential diagnosis is defined as the process of differentiating between the probability of a particular disease versus that of other diseases with similar symptoms that could possibly account for illness in a patient \cite{LAMBA20211}. Based on this concept, we propose a novel reward function $R(\mathbf{s}_t,\mathbf{s}_{t+1})$ to motivate our RL agent not only to examine symptoms with a high co-occurrence, but also to undertake a differential diagnosis process in the manner of real doctors. Inspired by HiePPO \cite{liu2021dialogue}, we first define $R(\mathbf{s}_t,\mathbf{s}_{t+1})$ as
\begin{equation}
R(\mathbf{s}_t,\mathbf{s}_{t+1})=\left\{
\begin{aligned}
R_{end}, &\ \ \    if\ dialogue \ ending\\  
R_{diff}, &\ \ \    otherwise\\  
    \end{aligned}  
\right.
\end{equation}
where $R_{end}$ is the long-term reward for the final diagnosis, and $R_{diff}$ is the short-term reward for evaluating the inquiry. When the dialogue ends, we set $R_{end}$ to $1$ if $\mathbf{y}_{pred}=\mathbf{y}_{true}$, otherwise to $-1$. 
$\mathbf{y}_{pred}$ is the diagnosis made by $\mathcal{D}$ and $\mathbf{y}_{true}$ is the true diagnosis. To evaluate the inquiry quality, we define a function $Diff(\mathbf{s}_t,x)$ to estimate the differential effect on the current state $\mathbf{s}_t$ if the patient has a certain symptom $x$:
\begin{equation}
\label{eq:diff}
    Diff(\mathbf{s}_t,x,\mathbf{s}'_t)=\lvert\ D_\textrm{KL}[\mathbf{y}\ ||\ \mathcal{D}(\mathbf{s}_t)]-D_\textrm{KL}[\mathbf{y}\ ||\ \mathcal{D}(\mathbf{s}'_t)]\ \rvert,
\end{equation}
where $\mathbf{s}'_t$ assumes that the patient has symptom $x$. The two KL divergences measure the difference between the true diagnosis result and the disease distribution predicted by the diagnosis model $\mathcal{D}$. Hence the absolute value of the difference between the two terms indicates the significance of $x$ given $\mathbf{y}$ and $\mathcal{D}$. Then, depending on whether or not $x$ is one of the implicit symptoms of the patient, we design two slightly different short-term rewards. If the patient truly experienced symptom $x$, $R_{diff}$ is simply $Diff(\mathbf{s}_t,x,\mathbf{s}_{t+1})$:
\begin{equation}
\label{eq:im}
R_{diff}=Diff(\mathbf{s}_t,x,\mathbf{s}_{t+1})
\end{equation}
since $\mathbf{s}'_t$ is equal to $\mathbf{s}_{t+1}$. Otherwise, for negative symptom $x$, we have
\begin{equation}
\label{eq:alpha}
R_{diff}=p(x|\mathbf{x}_O)\cdot Diff(\mathbf{s}_t,x,\mathbf{s}'_t)-\alpha,
\end{equation}
where the conditional probability $p(x|\mathbf{x}_O)\in(0,1)$ reduces the reward for unrelated symptoms based on observed data, which can be computed by the pretrained VAE as described in Sec.~\ref{sec:ga}. $\alpha$ is a constant penalty factor for lengthy interactions.

Our reward shaping function is more instructive than other approaches' greedy reward schemes \cite{liao2020task, peng2018refuel, liu2021dialogue}, which only consider positive symptoms. For example, HiePPO \cite{liu2021dialogue} simply returns a constant reward when a positive symptom is inquired about or when the rank of the true label's probability in the diagnosis result is moved up after an inquiry. The greedy scheme causes the sparse reward problem and offers a scalability barrier because positive symptoms account for a tiny proportion of all possible symptoms. Besides, our $R_{diff}$, which reflects the differential effect of symptoms, is more meaningful and fine-grained than the rank-based scheme of HiePPO. In conclusion, our reward shaping mechanism can allow more inquired symptoms, particularly negative ones with crucial differential effects, to receive more informative rewards, addressing the sparse reward problem in large feature spaces.

\begin{figure}
  \centering
  \includegraphics[width=\linewidth]{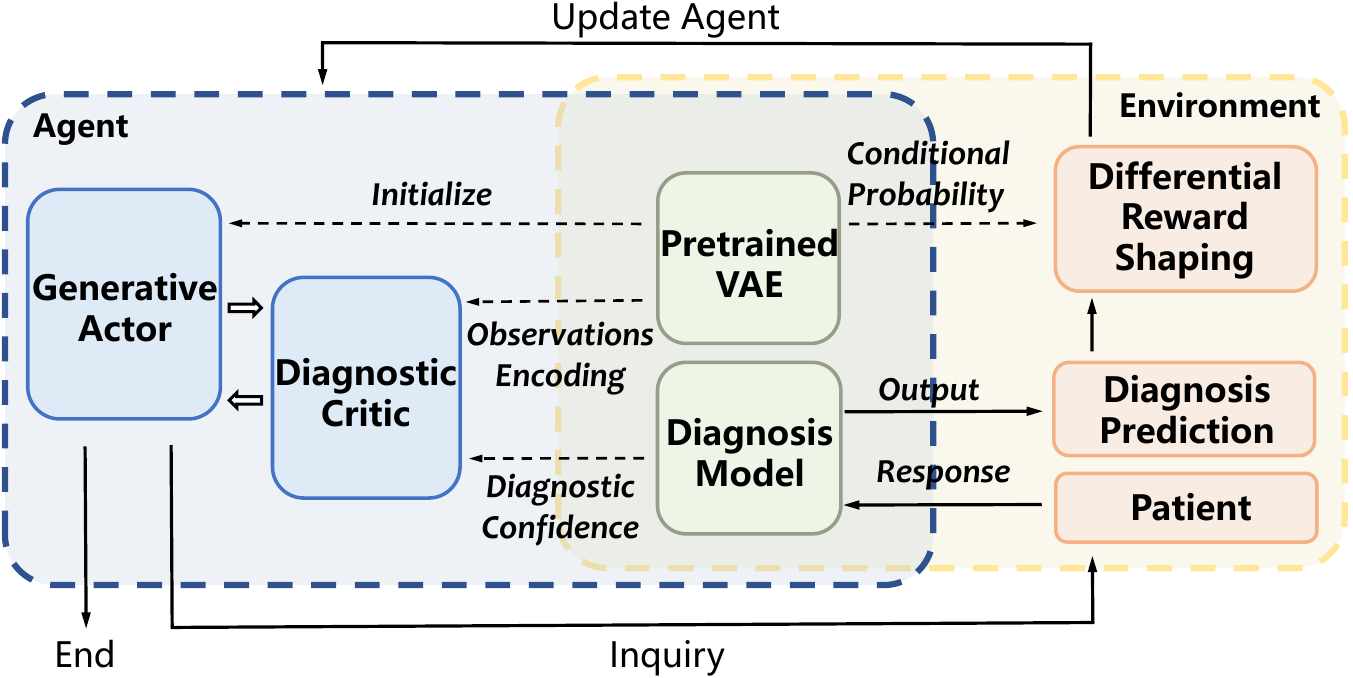}
  \caption{Overall flowchart of \mname.}
  \Description{The solid lines represent the \mname\ workflow with a single round of inquiry and diagnosis. When the actor decides to inquire about a specific symptom from the patient, the diagnosis model will produce an intermediate prediction based on the patient’s response. The differential reward shaping function then uses the prediction to compute the reward defined in Sec.~\ref{sec:r} and to update the agent. If the actor chooses to terminate the dialogue, the final diagnosis result would be the result of the last intermediate prediction. The dotted lines illustrate the effects of the generative and diagnosis models in \mname. Apart from producing a diagnosis prediction, the diagnosis model feeds the diagnostic confidence to the critic. The pretrained VAE initializes the actor’s parameters, and the conditional probability computed by the pretrained VAE is utilized for reward shaping. Additionally, the critic makes use of the pretrained VAE’s ability to deal with arbitrary partial observations.}
  \label{fig:flowchart}
\end{figure}

\subsection{Multi-Model-Fused Actor-Critic Reinforcement Learning}

We incorporated a generative actor network (Sec.~\ref{sec:ga}), a diagnostic critic network (Sec.~\ref{sec:diag}), a pretrained VAE, a supervised diagnosis model, and a differential reward shaping function to create a Multi-Model-Fused Actor-Critic reinforcement learning framework (\mname), as shown in Figure~\ref{fig:flowchart}. The structures of the actor and the critic networks are depicted in Figure~\ref{fig:ar}. The solid lines in Figure~\ref{fig:flowchart} represent the \mname\ workflow within a single round of inquiry and diagnosis. When the actor decides to inquire about a specific symptom from the patient, the diagnosis model will produce an intermediate prediction based on the patient’s response. The differential reward shaping function defined in Sec.~\ref{sec:r} then uses the prediction to compute the reward and update the agent. If the actor chooses to terminate the dialogue, the final diagnosis would be the result of the last intermediate prediction. The dotted lines in Figure~\ref{fig:flowchart} illustrate the effects of the generative and diagnosis models in \mname. Apart from producing a diagnosis prediction, the diagnosis model feeds the diagnostic confidence to the critic. The pretrained VAE initializes the actor’s parameters, and the conditional probability computed by the pretrained VAE is utilized for reward shaping. Additionally, the critic makes use of the pretrained VAE’s ability to deal with arbitrary partial observations.

In the training procedure, we begin by training the supervised diagnosis model that receives fully observed symptoms $\mathbf{x}$ as input. We drop a random fraction of the complete features $\mathbf{x}$ to obtain $\hat{\mathbf{x}}$, imitating arbitrary partial observations during the inquiry process. Then we pretrain a VAE model by the ELBO in Eqn.~\ref{eq:ELBO}, which outputs $p_\Theta(\hat{\mathbf{x}}|\mathbf{z})$ to recover the input $\hat{\mathbf{x}}$ as described in Sec.~\ref{sec:ga}. After that, we initialize the actor's encoder and decoder with the pretrained VAE's parameters and only the decoder is fine-tuned in the follow-up training. Next, we train the actor and critic with Proximal Policy Optimization (PPO) \cite{schulman2017proximal}. The critic network is trained using a mean squared error loss function: $L_{MSE}=\sum(V_{pred}-V_\pi(\mathbf{s}_t))^2$,
where $V_{pred}$ is the output of the critic and $V_\pi(\mathbf{s}_t)$ is calculated by Eqn.~\ref{eq:v}. According to the actor-critic algorithm \cite{mnih2016asynchronous} and the Generalized Advantage Estimator (GAE) \cite{schulman2015high}, let $\delta_t^V=r_t+\gamma V(\mathbf{s}_{t+1})-V(\mathbf{s}_t)$ and the advantage function optimized at timestep $t$ is
\begin{equation}
\label{eq:A}
    \hat{A}_t=\sum_{l=0}^T(\gamma\lambda)^l\delta_{t+l}^V\ ,
\end{equation}
where $T$ is the final step, $\gamma$ is the discount factor, and $\lambda$ is a hyperparameter of GAE. According to the clipped PPO objective function \cite{schulman2017proximal}, the goal of actor training is to maximize the objective function:
\begin{equation}
\label{eq:L}
    L(\theta)=\hat{\mathbb{E}}_t[\min(o_t(\theta) \hat{A}_t, clip(o_t(\theta), 1-\epsilon,1+\epsilon) \hat{A}_t)+\eta\cdot E(\theta)],
\end{equation}
where $o_t(\theta)=\frac{\pi_\theta(a_t|\mathbf{s}_t)}{\pi_{\theta_{old}}(a_t|\mathbf{s}_t)}$ is the probability ratio and $\theta$ is the parameters of the actor network. $E(\theta)=-\pi_\theta(a_t|\mathbf{s}_t)ln(\pi_\theta(a_t|\mathbf{s}_t))$ is the entropy of the policy $\pi_\theta$ to encourage more exploration and $\eta$ is a small weight. To avoid inquiring repeatedly, we set the probabilities of symptoms mentioned in the actor's output to zero. A maximum number of inquiries $N_T$ is also set to avoid lengthy interactions. Finally, when conducting the final diagnosis, we use the VAE to impute unobserved symptoms before feeding them to the diagnosis model to provide a more accurate diagnosis.

    
    





\section{Experiments}
\label{sec:exp}

\subsection{Datasets}

Obviously, due to patient privacy and security concerns, it is difficult to obtain large-size real-world medical data from the public domain for training and evaluation. Nevertheless, computational approaches should be evaluated for their generalizability and ability to handle large datasets. Hence synthetic datasets containing thousands of features are used to evaluate a model’s ability to handle large feature spaces. \mname\ shows excellent performance on both synthetic and real-world datasets. All datasets are publicly accessible and do not contain any personally identifiable information.

\subsubsection{Real-World Medical Dialogue Datasets}

The MuZhi medical dialogue dataset \cite{wei2018task} was collected from the dialogue data from the pediatric department of a Chinese online healthcare website (\url{https://muzhi.baidu.com/}). It includes 66 symptoms associated with 4 diseases: children's bronchitis, children's functional dyspepsia, infantile diarrhea infection, and upper respiratory infection. 

The Dxy medical dialogue dataset \cite{xu2019end} contains data from another popular Chinese online healthcare website (\url{https://dxy.com/}). It includes 41 symptoms for 5 diseases: allergic rhinitis, upper respiratory infection, pneumonia, children hand-foot-mouth disease, and pediatric diarrhea. All patients in the dataset are infants. These two datasets have been structured in the form of disease-symptoms pairs and split into a training set and a testing set. 




\subsubsection{Real-World Diabetes Dataset}

\citet{kachuee2018opportunistic} compiled a diabetes dataset from the NAtional Health and Nutrition Examination Survey (NAHNES) data \cite{nhr}. It consists of 45 features, such as demographic information, lab results (total cholesterol, triglyceride, etc.), examination data (weight, height, etc.), and questionnaire answers (smoking, alcohol, etc.). In contrast to the above symptom checking datasets, some of these features are continuous, and each feature is associated with a cost suggested by an expert in medical studies. The fasting glucose values were used to define three classes based on standard threshold values: normal, pre-diabetes, and diabetes. The dataset is randomly split into three subsets, including 15\% for testing, 15\% for validation, and the rest 70\% for training.

\subsubsection{Synthetic SymCAT Dataset}

SymCAT is a symptom-disease database \cite{symcat}. There is information about the symptoms associated with each disease, as well as the marginal probabilities associated with these symptoms. Using SymCAT, we construct a synthetic dataset in the manner described by \citet{peng2018refuel}. We randomly select a disease and its associated symptoms from among all disorders, and then conduct a Bernoulli trial on each extracted symptom based on the associated probability to create a symptom set. For example, if we take a sample of the disease "abscess of nose," we would observe the following symptoms: "cough" and "fever," which occur in 73\% and 62\% of cases, respectively. Then, based on these probabilities, we generate one data instance by sampling Bernoulli random variables. In this way, we created three diagnostic tasks with 200, 300, and 400 diseases, respectively, and for each task, we sampled $10^6$, $10^5$, and $10^4$ records, respectively for training, validation, and testing.

\subsubsection{Synthetic HPO Dataset}

The Human Phenotype Ontology (HPO) \cite{kohler2017human} provides a standardized vocabulary for phenotypic abnormalities associated with human diseases. The HPO’s initial domain of application was rare disorders. There are 123,724 HPO terms associated with rare diseases and 132,620 with common diseases. Following \citet{he2022bsoda}, we chose diseases and related symptoms from the public rare disease dictionary, along with their marginal probabilities, and used the aforementioned process for SymCAT to create synthetic datasets. Two diagnostic tasks containing 500 and 1,000 diseases were created for evaluation.

\begin{table*}
\caption{A case study for actual inquired symptoms on the Dxy dataset.}  
\label{case}

\begin{tabular}{cccc}  
\toprule  
Self-Report & Source & Inquired Symptoms & Diagnosis\\
\midrule  
\multirow{4}{*}{\shortstack{A 40-day Newborn, Sneezing,  \\ Cough, Spitting Bubbles, \\ Clear Nasal Discharge, \\ Difficult Breathing}} & BSODA & Convulsion, \textbf{Stuffy Nose}, Rubbing Nose, Coughing Up Phlegm & Allergic Rhinitis \\
\cmidrule(lr){2-4}    
& \mname & Loose Stools, Fever,  Coughing Up Phlegm, \textbf{Vomiting} & Pneumonia\\
\cmidrule(lr){2-4}  
& Doctor & \textbf{Vomiting}, \textbf{Stuffy Nose} & Pneumonia\\
\midrule  
\midrule  
\multirow{3.5}{*}{\shortstack{A 46-day Newborn, Sneezing,\\ Phlegm Sound, Stuffy Nose, \\ Cough, Choking milk}}  & BSODA & Rubbing Nose, Fever, \textbf{Difficult Breathing}, Runny Nose, Convulsion & Pneumonia\\
\cmidrule(lr){2-4}    
& \mname & Runny Nose, Fever, \textbf{Difficult Breathing} & Pneumonia\\
\cmidrule(lr){2-4}  
& Doctor & \textbf{Difficult Breathing} & Pneumonia\\
\bottomrule  
\end{tabular}  

\end{table*}

\subsection{Experimental Protocol}

\mname\ is compared with the following baselines:

\textbf{GAMP} \cite{xia2020generative} uses Generative Adversarial Network (GAN) to implement an RL agent for dialogue-based diagnosis, which has only been proven to be effective on MuZhi and Dxy.

\textbf{FlatPPO} \cite{liu2021dialogue} is a simplified version of HiePPO \cite{liu2021dialogue}, a customized hierarchical RL approach, which has been applied to MuZhi and Dxy. HiePPO can only take up to 90 diseases, which is too small to be included as our baseline for synthetic datasets.
    
\textbf{REFUEL} \cite{peng2018refuel} is a deep RL method that has been evaluated on medium sized synthetic datasets. 
    
\textbf{OL} \cite{kachuee2018opportunistic} is based on deep Q-learning and designed for cost-sensitive feature selection in datasets such as the diabetes dataset.
    
\textbf{BSODA} \cite{he2022bsoda} is a non-RL framework, which has shown state-of-the-art performance on a variety of datasets and the ability to deal with a large number of features. 
 

Patients’ self-reports, if available, are used at the beginning of the inquiring process. MuZhi and Dxy have already had self-reports. For the synthetic datasets, a positive symptom is chosen uniformly at random as a patient self-report. The diabetes dataset is evaluated without self-reports. 

Hyperparameters of \mname\ are set differently across datasets due to differences in the number of diseases and symptoms. $\alpha$ is set to 0 for MuZhi and Dxy, 0.1 for the diabetes dataset, the synthetic SymCAT dataset, and the synthetic HPO 500-disease task, and 0.15 for the synthetic HPO 1,000-disease task. The discount factor $\gamma$ in Eqn.~\ref{eq:A} is set to 0.95 for the synthetic SymCAT dataset and 0.9 otherwise. $\lambda$ in Eqn.~\ref{eq:A} is set to 0.9 and $\epsilon$ in Eqn.~\ref{eq:L} is set to 0.1. In Eqn.~\ref{eq:L}, $\eta$ is set to 0.001 for the synthetic HPO dataset and to 0.01 for all other datasets. We train the networks using the Adam optimizer \cite{kingma2014adam}. The learning rates of both the actor and critic are $5\times 10^{-5}$ and the batch size is 512. $\eta$ and the learning rates would linearly decay to one-tenth with iterations. The upper limit of dialogue rounds $N_T$ is set to 20. The evaluation metrics are diagnostic accuracy and the average number of rounds in dialogues. Our objective is to achieve high diagnostic accuracy while limiting the rounds of dialogues.

\subsection{Results}

\begin{table}\small

\caption{Performance of \mname\ and previous works on two medical dialogue datasets.}  
\label{tab:mzdxy}
\begin{center}
\begin{tabular}{ccccc}  
\toprule  
\multirow{2.5}{*}{Method} & \multicolumn{2}{c}{MuZhi} & \multicolumn{2}{c}{Dxy} \\
\cmidrule(lr){2-3} \cmidrule(lr){4-5}  
& Accuracy (\%) & \#Inquiries & Accuracy (\%) & \#Inquiries\\
\midrule
REFUEL & 71.8 & 6.20 & 75.7 & 5.27\\ 
GAMP & 73 & - & 76.9 & 2.68\\
FlatPPO & 73.2 & 6.32 & 74.6 & 3.31\\
BSODA & 73.1 & 8.80 & 80.2 & 7.49\\
\midrule
\mname\ & \textbf{73.9} & \textbf{4.47} & \textbf{82.7} & 6.08\\
\bottomrule  
\end{tabular}
\end{center}
\end{table}

\subsubsection{Real-World Medical Dialogue Datasets}

Table~\ref{tab:mzdxy} compares the results for \mname\ and previous works on the MuZhi and Dxy datasets. Since GAMP \cite{xia2020generative} did not disclose their source code or the average number of inquiries on MuZhi, we only present its self-reported accuracy on MuZhi. The results show that our method \mname\ outperformed the other techniques on MuZhi, achieving the highest accuracy and requiring the fewest inquiry rounds. Our method is also the most accurate on Dxy, although it generates more queries than REFUEL \cite{peng2018refuel}, GAMP \cite{xia2020generative}, and FlatPPO \cite{liu2021dialogue}. Overall, \mname\ remains the best method because its accuracy is at least 5.8 points higher than the other methods. Additionally, the number of inquiries generated by \mname\ is less than that by the previous SOTA method, BSODA \cite{he2022bsoda}.

Using the Dxy dataset, we conducted a case study to investigate the actual inquired symptoms of BSODA, \mname\ and the real doctor. The results are summarized in Table~\ref{case} with the inquired implicit symptoms highlighted in bold. The upper case is a 40-day infant who had \emph{Pneumonia} but was easily misdiagnosed as having \emph{Allergic Rhinitis} due to the presence of both nasal symptoms and infantile pneumonia signs, such as spitting bubbles and difficult breathing. It is worth noting that infant may exhibit no symptoms of pneumonia infection. Alternatively, they may vomit, have a fever and cough, or have diarrhea. These are not typical symptoms of \emph{Allergic Rhinitis}. As a result, \mname\ inquired about \emph{Loose Stools}, \emph{Fever} and \emph{Vomiting} in order to differentiate the two illnesses. While BSODA tended to inquire more about nasal symptoms based on the initial "\emph{Clear Nasal Discharge}", resulting in an erroneous misdiagnosis. It demonstrates that, using differential reward shaping, \mname\ makes inquiries based on the differential effect rather than depending on co-occurrence of symptoms such as BSODA.

Another case involves a 46-day old infant who is suffering from \emph{Pneumonia}. The key implicit symptom is \emph{Difficult Breathing}. Although both BSODA and \mname\ inquired about \emph{Difficult Breathing}, we observe that \mname\ required fewer inquiries before terminating the interaction and providing a confident diagnosis, which demonstrates its efficiency.

\begin{table}\small
\caption{Performance of OL \cite{kachuee2018opportunistic}, BSODA \cite{he2022bsoda} and \mname\ on the diabetes dataset.}  
\label{tab:db}
\begin{center}
\begin{tabular}{ccc}  
\toprule  
Method & Accuracy (\%) & Costs\\
\midrule
OL & 81.66 & 63.99\\
BSODA & 86.11 & 23.49\\
\midrule
\mname\ & \textbf{87.18} & \textbf{20.31}\\

\bottomrule  
\end{tabular}
\end{center}
\end{table}

\subsubsection{Real-World Diabetes Dataset}

Apart from symptom checking tasks, Table~\ref{tab:db} shows an evaluation using the real-world diabetes dataset that has hybrid types of features, such as laboratory test results and questionnaire answers. The total costs for all 45 features are 185. It should be noted that symptom checking baselines such as REFUEL \cite{peng2018refuel} and FlatPPO \cite{liu2021dialogue} cannot be directly applied to these continuous features since they only construct reward functions for implicit symptoms instead of continuous features as in the diabetes dataset. These baselines are incapable of differentiating the significance of various continuous features in a patient. For \mname, however, these feature types have no effect on the differential function in Eqn.~\ref{eq:diff}, so it could handle these features only with the reward in Eqn.~\ref{eq:im}. We observe that both BSODA \cite{he2022bsoda} and \mname\ outperform OL \cite{kachuee2018opportunistic} in terms of accuracy while having a lower acquisition cost, with \mname\ being the superior method. Thus, in addition to performing symptom checking, \mname\ can handle online diagnosis problems involving hybrid data types with the help of the differential reward shaping function.

\begin{table}\small
\caption{Performance of REFUEL \cite{peng2018refuel}, BSODA \cite{he2022bsoda} and \mname\ on synthetic datasets.}  
\label{acc}

\begin{tabular}{cccccc}  
\toprule  
Task&Method&Top1&Top3&Top5&\#Inquiries\\
\midrule 
\multirow{3}{*}{\shortstack{SymCAT\\(200 Diseases\\328 Symptoms)}} & REFUEL & 53.76 & 73.12 & 79.53 & 8.24\\
& BSODA & 55.65 & 80.71 & 89.32 & 12.02\\
& \mname\ & \textbf{59.00} & \textbf{84.66} & \textbf{92.45} & 12.54\\
\midrule
\multirow{3}{*}{\shortstack{SymCAT\\(300 diseases\\349 Symptoms)}} & REFUEL & 47.65 & 66.22 & 71.79 & 8.39\\
& BSODA & 48.23 & 73.82 & 84.21 & 13.10\\
& \mname\ & \textbf{51.91} & \textbf{78.24} & \textbf{87.55} & 13.80\\
\midrule
\multirow{3}{*}{\shortstack{SymCAT\\(400 diseases\\355 Symptoms)}} & REFUEL & 43.01 & 59.65 & 68.89 & 8.92\\
& BSODA & 44.63 & 69.22 & 79.54 & 14.42\\
& \mname\ & \textbf{45.80} & \textbf{72.14} & \textbf{82.43} & 14.73\\
\midrule
\multirow{3}{*}{\shortstack{HPO\\(\textbf{500} diseases\\\textbf{1901} Symptoms)}} & REFUEL & 64.33  & 73.14  & 75.34  & 8.09\\
& BSODA & 76.23 & 84.17 & 86.99 & 5.34\\
& \mname\ & \textbf{88.39} & \textbf{93.96} & \textbf{94.96} & \textbf{4.77}\\
\midrule
\multirow{3}{*}{\shortstack{HPO\\(\textbf{1000} diseases\\\textbf{3599} Symptoms)}} & REFUEL & 40.08  & 62.67  & 67.42  & 14.19\\
& BSODA & 67.03 & 75.94 & 79.27 & 10.62\\
& \mname\ & 63.04 & \textbf{77.99} & \textbf{81.43} & \textbf{10.61}\\
\bottomrule  
\end{tabular}  

\end{table}

\subsubsection{Synthetic SymCAT Dataset}

For the three diagnostic tasks with 200, 300, and 400 diseases, among RL baselines, only REFUEL \cite{peng2018refuel} was found to be effective in these tasks with a medium search space encompassing hundreds of symptoms. The results in Table~\ref{acc} demonstrate that, on average, BSODA \cite{he2022bsoda} and \mname\ conduct more inquiries than REFUEL, but resulting in significantly higher accuracy. REFUEL converges more quickly but produce a less accurate prediction. As BSODA claimed \cite{he2022bsoda}, with a few additional inquiries, it may be able to provide a significantly improved prediction, which is critical in the medical domain. Compared to BSODA, \mname\ makes nearly the same number of inquiries but yields higher accuracy, resulting in the best performance on these tasks.

\subsubsection{Synthetic HPO Dataset} Two synthetic HPO tasks containing 500 and 1,000 diseases, respectively, are significantly larger than the SymCAT data. The results in Table~\ref{acc} show that both \mname\ and BSODA \cite{he2022bsoda} outperform REFUEL \cite{peng2018refuel} by a large margin while making fewer inquiries. It demonstrates that the traditional RL baseline struggles in a large feature space. In comparison, our RL approach \mname\ demonstrates exceptional scalability, outperforming BSODA in almost every statistical measure except one.

\subsubsection{Reaction Time} It should be noted that \mname\ has a clear advantage over BSODA in terms of reaction time. BSODA conducts Monte Carlo sampling for every candidate symptom to estimate their reward, resulting in \emph{quadratic time complexity}. Its time cost increases as the numbers of symptoms and samples increase, whereas \mname\ \emph{always requires only one forward propagation of the actor network}. Using an NVIDIA GeForce GTX 1080Ti GPU, BSODA took an average of \textbf{0.95 seconds} to determine an action in the 1,000-disease task, while \mname\ took only \textbf{0.02 seconds}. Thus, even with only a slightly lower performance in the Top 1 category of the 1,000-disease task, \mname\ remains the most competitive and efficient, and it is more suitable for online applications.

\subsection{Ablation Study}

\begin{figure}
  \centering
  \includegraphics[width=0.9\linewidth]{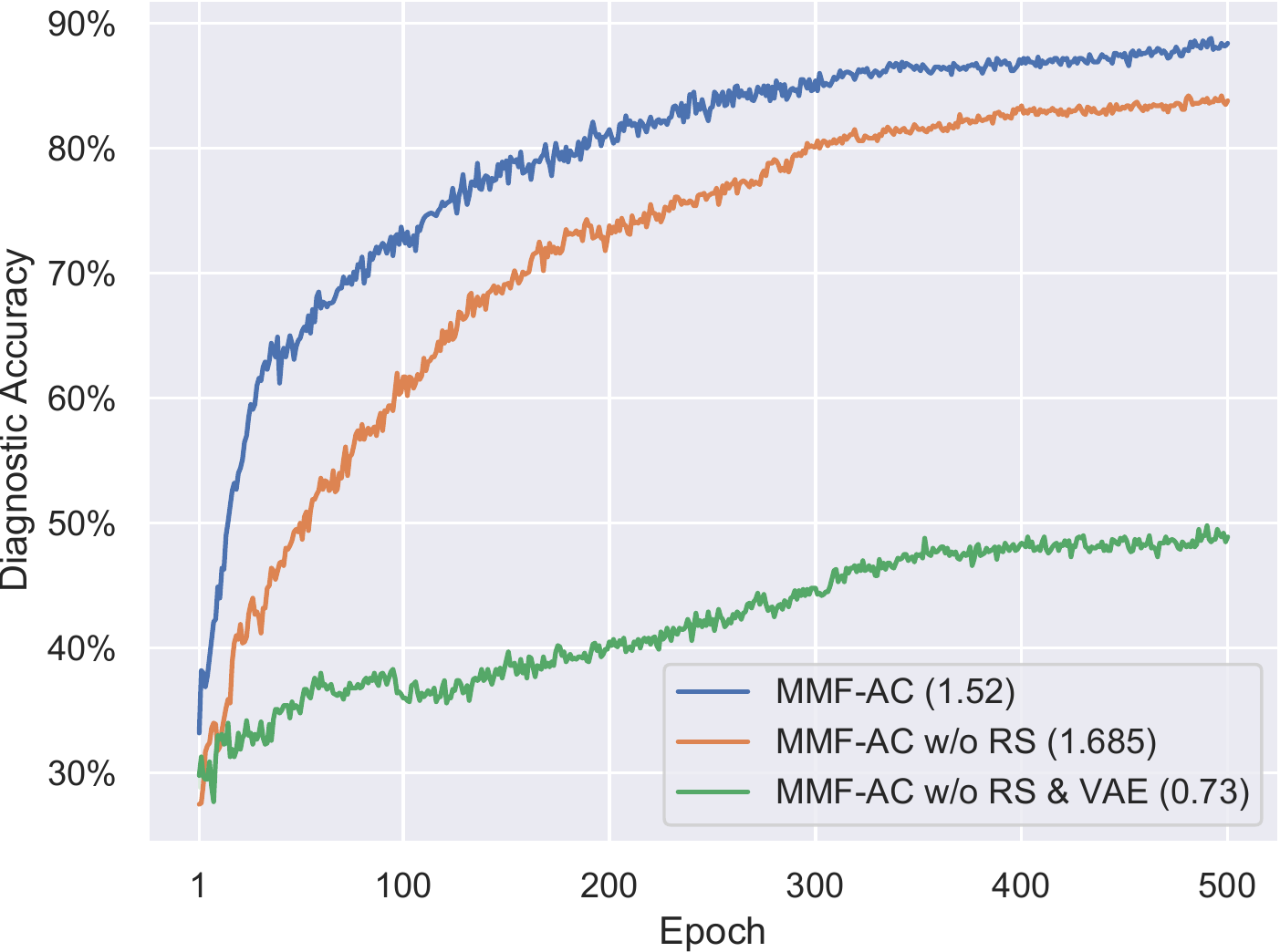}
  \caption{Learning curves of the three \mname\ settings on the synthetic HPO dataset.}
  \Description{As the number of epochs increases, \mname\ without the handling of partial observations by VAE finally converges to a very bad policy. If the differential reward shaping is not applied, \mname\ could inquire more implicit symptoms based on the reward of co-occurrence but encounters the difficulties of the further improvement.}
  \label{fig:ab}
\end{figure}

We performed an ablation study to investigate the effects of the VAE and the differential reward shaping mechanism using the synthetic HPO 500-disease task. In Figure~\ref{fig:ab}, "\mname\ w/o RS" denotes that we omit the differential reward shaping function and instead set $R_{diff}$ simply to 0.25 when inquiring about an implicit symptom. "\mname\ w/o RS and VAE" denotes that we omit the effects of both the reward shaping and the VAE. The generative actor is then a simple MLP, and $\mathbf{z}_t$ encoded by the VAE in the critic is replaced by the observed state $\mathbf{s}_t$. The number in the parenthesis denotes the average number of implicit symptoms queried. We observe that, with an increasing number of epochs, \mname\ without partial observations handled by the VAE converges slowly in the initial stage and eventually to a horrible policy, highlighting the critical role of the generative model in large search spaces. Without differential reward shaping, \mname\ is able to inquire more implicit symptoms based on symptom co-occurrence, but performs poorly due to the omission of the differential effect of symptoms. It indicates that the differential effect of symptoms is more important than their co-occurrence in making an accurate diagnosis.

\section{Conclusion}

We offer \mname, a novel Multi-Model-Fused Actor-Critic reinforcement learning approach for online disease diagnosis that is scalable to large feature spaces. It incorporates a generative model and a supervised classification model into an actor-critic RL algorithm. We also develop a novel reward shaping mechanism inspired by differential diagnosis. The experimental results on both real-world and synthetic datasets demonstrate the effectiveness of our approach, which outperforms prior RL and non-RL methods in their respective settings \cite{peng2018refuel,kachuee2018opportunistic,xia2020generative,liu2021dialogue,he2022bsoda}. The scalability to large feature spaces and generalizability for handling hybrid feature types make our approach suitable for online applications.

\begin{acks}

This work is supported by the National Key R\&D Program of China (2021YFF1201303 and 2019YFB1404804), National Natural Science Foundation of China (grants 61872218), Guoqiang Institute of Tsinghua University, Tsinghua University Initiative Scientific Research Program, and Beijing National Research Center for Information Science and Technology (BNRist). The funders had no roles in study design, data collection and analysis, the decision to publish, and preparation of the manuscript.
\end{acks}

\bibliographystyle{ACM-Reference-Format}
\bibliography{1}

\end{document}